\documentclass[11pt,a4paper]{article}
\usepackage[hyperref]{acl2018}
\usepackage{times}
\usepackage{latexsym}
\usepackage{pgfplots}
\usepackage{url}
\usepackage{amssymb}

\aclfinalcopy % Uncomment this line for the final submission
\title{A Hierarchical Approach to Neural Context-Aware Modeling}

\author{Patrick Huber, Jan Niehues, Alex Waibel \\
	Institute for Anthropomatics and Robotics \\
  	Karlsruhe Institute of Technology \\
	Karlsruhe, Germany \\
 	{\tt uhejt{@}student.kit.edu},  {\tt  \{jan.niehues, alex.waibel\}{@}kit.edu}
  \\}

\date{}

\begin{document}
\maketitle
\begin{abstract}
We present a new recurrent neural network topology to enhance state-of-the-art machine learning systems by incorporating a broader context. Our approach overcomes recent limitations with extended narratives through a multi-layered computational approach to generate an abstract context representation. Therefore, the developed system captures the narrative on word-level, sentence-level, and context-level. Through the hierarchical set-up, our proposed model summarizes the most salient information on each level and creates an abstract representation of the extended context. We subsequently use this representation to enhance neural language processing systems on the task of semantic error detection. 

To show the potential of the newly introduced topology, we compare the approach against a context-agnostic set-up including a standard neural language model and a supervised binary classification network. The performance measures on the error detection task show the advantage of the hierarchical context-aware topologies, improving the baseline by 12.75\% relative for unsupervised models and 20.37\% relative for supervised models.
\end{abstract}

\section{Introduction}
In the research field of natural language processing (NLP), most state-of-the-art algorithms and topologies for translation and transcription tasks are based exclusively on the local context of a single sentence. This restriction is reasonable when working with techniques such as n-gram models. Modern machine-learning approaches, for example neural networks, have become far more powerful and thus lifted the restriction of the models to a local context. 

Even though many systems shifted from classical approaches towards neural design patterns, the scope of these approaches is still limited to a local context. Through the sequential nature of NLP systems, considering a longer context can benefit the performance in many sub-domains. 

For instance in the domain of neural machine translation (NMT), state-of-the-art approaches are mostly utilizing the advantages of asynchronous encoder-decoder neural network designs. Extending these models with a context component has the potential to enhance the overall system and increase the reliability of the translation. 

Another example where the context-aware extension can add direct value to the computation is the area of automatic speech recognition (ASR). As a part of the fundamental equation of speech recognition, the language model can be improved by extending the employed context.

\section{Related Work}
\citet{translation_hierarchical} describe a hierarchical approach for NMT systems that takes an extended context into account. The described methodology in the paper summarizes the context in two computational stages: on sentence- and document-level. With this attempt to capture the context, the work by \citeauthor{translation_hierarchical} is closely related to our work. Nevertheless, the main goal of the paper is to avoid translation errors through consistent translations for equal words. In contrast to the research by \citeauthor{translation_hierarchical}, we additionally pre-train the sentence embeddings to reduce the required amount of training data.

In the paper \textit{Character-Level Language Modeling with hierarchical Recurrent Neural Networks} published by \citet{hwang_et_al_16} hierarchical embeddings are used to create abstract representations on word and sentence level to enhance recent character-based approaches for speech recognition tasks. Even though \citeauthor{hwang_et_al_16} utilize a hierarchical approach to summarize information on multiple layers, the scope differs significantly from our work, as we try to create an abstract representation of the narrative rather than only summarizing up to a sentence-level.

Further related work has been conducted by \citet{chung_et_al_16}, analyzing the capability of hierarchical neural networks to capture temporal dependencies within sequences. As a result of their work, they showed empirical evidence that this type of models can actually capture the temporal structure. \citeauthor{chung_et_al_16} also limit their system to lower-level features on word and sentence-level, not considering the context on document-level.

Related work within the area of question answering is conducted by \citet{nlp_example_related_iyyer} showing that recursive neural networks can be combined with a parse tree \citep{dependency_parse_tree} and a multinomial logistic regression classifier to achieve good results on trivia-like tasks by extracting the essential information from the context. The task introduced by \citeauthor{nlp_example_related_iyyer} varies from this work, as it targets a question-answering problem based on a short paragraph rather than semantic error detection on long narratives. Additionally, it utilizes a heterogeneous methodology instead of an end-to-end connectionist learning approach. 

The recently published work by \citet{iyyer2017search} is related to our work, tackling the topic of neural semantic parsing. The described approach maps natural language to a formal logic with the goal to answer trivia-like questions by utilizing weakly supervised methodologies. Even though the approach by \citeauthor{iyyer2017search} also targets the broader domain of natural language processing, the exact task as well as the used methodologies discriminate the work from ours.

Other related work in the area of question answering is conducted by \citet{kumar_related}, introducing a new neural network architecture to retrieve an answer from a long input sequence and a question statement. The described approach differs from our work, as it focuses on a neural network based framework creation rather than a methodology to capture an extended narrative. 

\section{Motivation}
\label{motivation}
Recent neural language models predict the next word in a sentence according to the syntactical fit into the text. As these models only take a single sentence at a time into account, standard neural languages models are not capable to infer a broader context. In many real-life applications, such as document translation, transcription of natural language and chatbots, the semantic coherence plays a crucial role to advance the quality of the overall system and avoid substantial mistakes. Imagine a transcription system assembling the following English sentence \footnote{Example taken from the TEDTalk \textit{Wireless data from every light bulb} by Harald Haas}:\\

\textit{And it's this importance why I decided to look into the issues that this technology has, because it's so fundamental to our \textbf{countries}.}\\

Running a sentence-based neural language model on this sentence will return a high likelihood, due to the syntactical coherence of the text passage. The model cannot find the semantic out-of-context token \textit{countries}, which should be transcribed as \textit{lives}. Considering the broader context, the semantic error in the sentence becomes evident, as it does not fit into the overall context. \\

\textit{We use it everyday. We use it in our everyday lives now, in our private lives, in our business lives. [...]. And it's this importance why I decided to look into the issues that this technology has, because it's so fundamental to our \textbf{countries}.}\\

Taking the extended narrative into account, the last word in the original sentence emerges to be out-of-context.

To infer this semantic mismatch, the computational system needs to take more than just the sentence itself into account. This ability to include the former context is a complex task to integrate into computational systems. In comparison to that, humans have the innate capability to easily take long contexts into account when listening or reading. We thereby do not keep every single word within the context in our limited short-term memory, but hierarchically structure the received information and only keep a chunked representation of the narrative \citep{newell1958elements, chomsky2002syntactic}. The human approach of hierarchical information retrieval allows us to group word-level units and acquire abstract representations of the world \citep{lashley1951problem, rosenbaum1983hierarchical}. This natural paradigm efficiently summarizes the most salient information of the context in an abstract representation. To overcome the recent limitation of computational models and achieve more human-like context understanding for neural networks, we determine whether a hierarchical approach, derived from the human information retrieval paradigm, can enhance the semantic error detection within contextual text passages compared to systems solely relying on a local context. 

\section{Task}
\label{task}
Within the area of text understanding, the evaluation task is defined as the detection of semantic errors in contextual text passages. To achieve that, artificially inserted out-of-context error tokens are uniformly distributed over a dataset. The respective replacement procedure is implemented in five-stages:\\

\noindent \textbf{(1) Dataset Filtering} to enhance the data quality\\
 
\noindent \textbf{(2) Part-of-Speech Tagging} substituting only nouns, as described superior by \citet{Paperno-et-al-16}, showing that the context is especially critical to nouns\\
 
\noindent \textbf{(3) Candidate Selection} to filter for nouns with strong contextual coherence\\
 
\noindent \textbf{(4) Appearance Window} approach to find syntactically suited replacement tokens by the word-count on the dataset\\
 
\noindent \textbf{(5) Tense and Grammatical Number} are taken into account within the appearance window to determine the most suitable substitution\\
 
The semantic word replacements within this task are at random positions w\textsubscript{r} in the series of word-tokens $W =(w\textsubscript{1},w\textsubscript{2},...,w\textsubscript{n})$ with $w\textsubscript{r} \in W$. The task can therefore be interpreted as a binary sequence labeling problem defined by the input sequence $W$ and the output sequence $L\textsubscript{ooc} = (l\textsubscript{1}, l\textsubscript{2}, ..., l\textsubscript{n})$ with $l\textsubscript{r} \in L\textsubscript{ooc}$, representing the labels of the words in $W$. The labels $L\textsubscript{ooc}$ separate the two classes $\{0, 1\}$, defining valid-context tokens and out-of-context tokens. The task definition does not provide any information about the position of the modifications nor give any insight about the total number of out-of-context substitutions in the data. With this definition, every word $w\textsubscript{r} \in W$ needs to be assessed against every other word $w\textsubscript{q} \in W$ ($r \neq q$) in the sequence in order to find the out-of-context tokens.

To assess the performance of our new model topology, a ground-truth dataset is created by extending the 2016 TEDTalk dataset \citep{wit3} with semantic errors at arbitrary positions and an error-insertion rate of 10 replacements per TEDTalk. We use the 2000 individual TEDTalks (with combined 1millions words) for both, the pre-training as well as the final training process. For the translation system, the 2016 synchronous English-German TEDTalk corpus is utilized. The three mutually exclusive datasets (training, development and test) are separated with a classical 60-20-20 split, which is used to randomly distribute the data. An example for a replacement on the training set is given below.\\

\noindent Modified sentence: \textit{So many artists, so many different explanations, but my explanation for engineering is very simple.}\\

\noindent Ground truth: engineering $\rightarrow$ performance\\

\noindent (Marina Abramovic, TED2015)\\

Through the automated modification process the whole dataset is altered, allowing unsupervised and supervised models to be trained on the data and evaluated against the task.

\section{Model Topology}
Our newly proposed neural network model topology introduces a context component, based on the recurrent encoder-decoder neural network design described by \citet{seq2seq_with_nn}. The asynchronous model architecture thereby allows the necessary two-staged computation, similar to state-of-the-art translation systems. 

\subsection{Context Representation}
\label{CR}
The encoder component of the model feeds the context into the network. A standard encoder, as proposed in the original paper \citep{seq2seq_with_nn}, processes one word at every time step. Simply feeding the extended context into the model this way is problematic, as the amount of time steps in the encoder drastically increases for long narratives with multiple sentences or paragraphs. 

To effectively augment the decoder component of the model with a narrative, the information of the context is hierarchically summarized. This way, an abstract representation of the context is generated, which can be fed into the fixed-size context vector of the encoder-decoder model. The narrative is summarized on three levels:\\
\\
\\
\noindent \textbf{(1) Word-Level}\\
The first level of abstraction is achieved through a word-embedding layer. This way, the network input  $W = \{w\textsubscript{1}, w\textsubscript{2}, ..., w\textsubscript{n}\}$ is transformed into real valued vectors by applying the parameterized function $E(W)$:
\begin{equation}
E(W) : W \rightarrow \mathbb{R}\textsuperscript{n $\times$ d}
\end{equation}
The word embedding function $E(W)$ is implemented as a lookup matrix of dimension $|V| \times d$, with $|V|$ as the vocabulary size and $d$ representing the word embedding dimension. The function transforms the sparse network input of size $|V|$ into a $d$ dimensional real valued vector with $d \ll |V|$. During the training process, the word embeddings learn to abstract from specific words and create a first synopsis on word-level \citep{bengio}.\\

\noindent \textbf{(2) Sentence-Level}\\
On the second layer of abstraction, the sentences $S = \{s\textsubscript{1}, s\textsubscript{2}, ..., s\textsubscript{m}\}$ in the dataset are embedded. To generate this representation, the word-embeddings $E(W)$ are fed sentence-wise into a recurrent neural network \citep{rnn_basics}. For a single sentence $s\textsubscript{j}$ within the context $C$, the sentence-level input is defined as $\{E(w\textsubscript{1, j}), E(w\textsubscript{2, j}), ..., E(w\textsubscript{n, j})\}$. During training, the network learns to store the important information of the sentence within the recurrent neural units defined by 
\begin{equation}
h\textsubscript{i, j} = f(h\textsubscript{i-1, j}, w\textsubscript{i, j})
\end{equation}
with $h\textsubscript{i, j}$ representing the hidden state of the network at time step $i$ of sentence $j$. $f(\cdot)$ is the activation function of the neural network layer. Through the sequential nature of the recurrent neural network, the final hidden state $h\textsubscript{n\textsubscript{j}, j}$ of the network contains an abstract representation of sentence $j$. Therefore, the final hidden state is used as the sentence representation.
\begin{equation}
H(s\textsubscript{j}) \equiv h\textsubscript{n, j}
\end{equation}

\begin{figure}[!ht]
\centering
\includegraphics[width=0.45\textwidth]{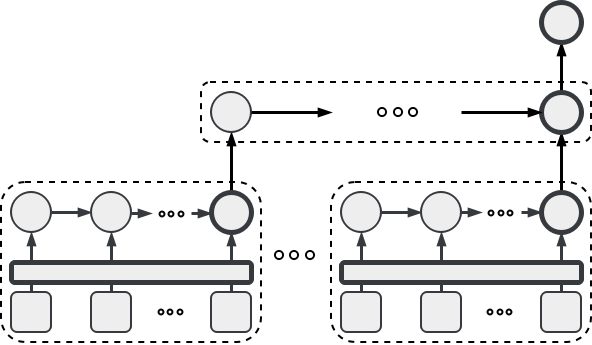}
\caption{Hierarchical Context Representation - Inputs are displayed as squares; recurrent neural network units are indicated by circles. The three layers of abstraction and the context vector are bolded.}
\label{fig:complete_model}
\end{figure}

\noindent \textbf{(3) Context-Level}\\
The sentence-wise content of the narrative $C$ is subsequently fed into the recurrent neural network representing the encoder component of the major encoder-decoder network. The context representation utilizes the summarized sentence representations $H(S) = \{H(s\textsubscript{1}), H(s\textsubscript{2}), ..., H(s\textsubscript{m})\}$ to add the final layer of abstraction to the model topology. The final hidden layer computed by 
\begin{equation}
h\textsubscript{m} = f(h\textsubscript{m-1}, s\textsubscript{m})
\end{equation}
is used as the context $C$ of the system.
\begin{equation}
C \equiv h\textsubscript{m} 
\end{equation}
The hierarchically summarized context $C$ is used as the context vector to initialize the decoder, containing the content of the complete narrative. (See figure \ref{fig:complete_model})\\

With the described three-layered approach we propose a new neural network architecture that learns to create an abstract representation of the context $C$. The model topology thereby allows to:\\ (a) Replace synonyms and closely related terms on a word level to abstract from the exact terminology used in the text.\\ (b) Determine the semantic of a sentence independent of the sentence structure.\\ (c) Create an abstract representation of the context, which models the human principle of the \textit{mindset}, as motivated in section \ref{motivation}.

To keep the computational efforts reasonable, the sentence embeddings are pre-trained on the same corpus as the final encoder-decoder model and are used for the presented evaluation of the system. The pre-trained sentence embeddings are described in more detail in section \ref{training}.

\subsection{Sentence Prediction}
The context vector of the encoder recurrent neural network initializes the decoder component. For our task, three different decoder topologies are generated:\\

\noindent \textbf{(1) Contextual Language Model}\\
The first decoder topology utilizes an unsupervised neural language model to synchronously predict the next word within the current sentence. Augmented with the context vector of the encoder computation, the model assigns every word in the vocabulary $V$ a probability, indicating the likelihood of the word to succeed the current word in the sentence. Through the encoder-decoder topology, the network jointly learns to enhance the context representation and the network output during the training.\\

\noindent \textbf{(2) Contextual Attention-based Language Model}\\
The attention based topology extends the neural language model with an additional attention mechanism derived from the paper by \citet{attention}. The attention component is introduced to remove the restriction of the standard encoder-decoder model to a fixed-size context vector and allow the decoder to dynamically focus on the encoder states.\\

\noindent \textbf{(3) Contextual Binary Classification Model}\\
As the last decoder topology, a supervised network is evaluated on the task. This model labels the input sentence through a binary classification on a word base and directly outputs the probability of the token within the context.

\section{Training}
\label{training}
To keep the model topologies comparable during the training and evaluation phase, the hyper-parameters of the models are identical for all instances. This includes the Adam Optimizer as described by \citet{adam}, the cross entropy cost function and the Long short-term memory (LSTM) cell architecture for the recurrent networks, as proven more efficient by \citet{lstm}. The networks are limited to 512 neurons per recurrent layer and a single recurrent layer per network. The mini batch-size for the training of the models is set to 100 data points. 

The attention based models use the Bahdanau attention mechanism \citep{attention}. 

Applied learning rates are derived from the original paper introducing the Adam Optimizer \citep{adam}. A vocabulary size $|V|$ of the most frequent $30,000$ words and a maximum sentence length $|S|$ of $50$ word-tokens are chosen. 

As mentioned in section \ref{CR}, the sentence embeddings are pre-trained for this evaluation in order to keep the computational efforts low. Furthermore, for a given sentence $s\textsubscript{j}$, the context $C\textsubscript{j}$ is defined as 
\begin{equation}
C\textsubscript{j} \equiv h\textsubscript{j-1} \
\end{equation}
where $h\textsubscript{j-1}$ is the last hidden state of the limited sentence representation $H(S\textsubscript{j,p}) = {H(s\textsubscript{j-p}), ..., H(s\textsubscript{j-1})}$ with $p$ set to $10$. The most salient information on a sentence-level is summarized by the following two approaches:\\

\noindent \textbf{(1) Final State of Neural Language Model}\\
The first approach to summarize the context on a sentence-level is realized by utilizing a neural language model. Thereby, the sentence $s\textsubscript{j}$ is fed into the network word-by-word and the final network state is extracted. The unsupervised neural language model is trained on the same TEDTalk training set as the final context-aware encoder-decoder model. This sentence-level abstraction exactly resembles the summarization of the network topology described in section \ref{CR}\\

\noindent \textbf{(2) Context Vector of Neural Machine Translation}\\
The second approach to abstract on the sentence-level is though the context vector of a NMT encoder-decoder model. The model is trained to translate bilingual TEDTalks from English into German. The training-, development- and test-set are aligned with the monolingual data split.\\

\begin{figure}[!ht]
\centering
\includegraphics[width=0.45\textwidth]{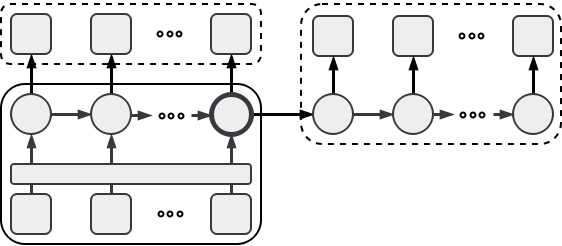}
\caption{Pre-Trained Sentence Representations - The solid box contains the network to compute the sentence representation; the dashed boxes enclose the additional network components during training.}
\label{fig:sentence_rep}
\end{figure}

Figure \ref{fig:sentence_rep} shows the network topologies of the pre-trained sentence embeddings. 

\section{Evaluation}
The evaluation of the hierarchical context-aware model topology is based on the task described in section \ref{task}. Within the evaluation, two common performance measures are applied on the newly introduced model, as well as the baseline. 

\subsection{Baseline}
To evaluate the impact of the narrative, the novel hierarchical context-aware model topology is assessed against sentence-based baseline models resembling the decoder part of the network without the context component. This way, the effect of the narrative can be individually tested and evaluated. The two baseline models are a neural language model and a neural binary classification model.

\subsection{Scoring}
In order to evaluate the performance of the presented models, the F-score is assessed on the task defined in chapter \ref{task}. As the total amount and the position of the semantic error tokens are not disclosed within the task, every word in the dataset $D$ is assessed according to its fit in the context. The scores are subsequently sorted and the least likely $N\textsubscript{ooc}$ words are classified as out-of-context tokens. As the number $N\textsubscript{ooc}$ of out-of-context tokens is not known, a suitable threshold is defined to separate the binary classes. To determine the best split, the F-score is calculated for every threshold $t\textsubscript{N\textsubscript{ooc}} = \{1 .. N\}$ with $ 1 \le N \le |D|$. For unsupervised models, the perplexity measure is additionally applied to show the systems' performance on the general word prediction task.

\subsection{Results}
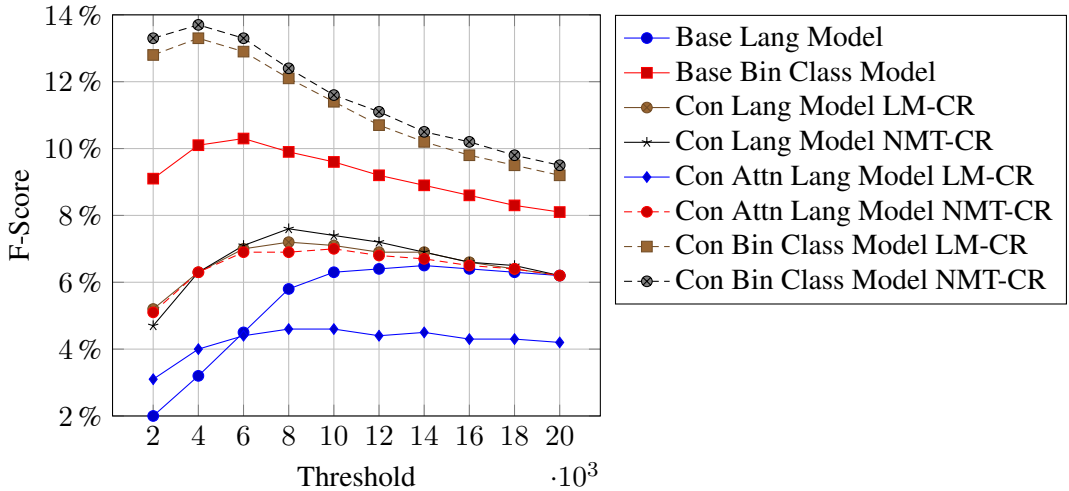
\begin{figure*}[!h]
\centering
\begin{tikzpicture}
\begin{axis}[
    legend pos= outer north east,
    legend cell align={left},
    width=0.5\textwidth,
    xlabel={Threshold},
    ylabel={F-Score},
    ymax = 14,
    ymin=2,
    xtick={2e3,4e3,6e3,8e3,10e3,12e3,14e3,16e3,18e3,20e3},
    scaled x ticks={base 10:-3},
    ytick={2,4,6,8,10,12,14},
    yticklabel=\pgfmathparse{\tick}\pgfmathprintnumber{\pgfmathresult}\,\%,    
    ymajorgrids=true,
    xmajorgrids=true,
]
\addplot
    coordinates {(2000,2.0)(4000,3.2)(6000,4.5)(8000,5.8)(10000,6.3)(12000,6.4)(14000,6.5)(16000,6.4)(18000,6.3)(20000,6.2)};
    \addlegendentry{Base Lang Model}
    
\addplot
    coordinates {(2000,9.1)(4000,10.1)(6000,10.3)(8000,9.9)(10000,9.6)(12000,9.2)(14000,8.9)(16000,8.6)(18000,8.3)(20000,8.1)};
    \addlegendentry{Base Bin Class Model}
    
\addplot
    coordinates {(2000,5.2)(4000,6.3)(6000,7)(8000,7.2)(10000,7.1)(12000,6.9)(14000,6.9)(16000,6.6)(18000,6.4)(20000,6.2)};
    \addlegendentry{Con Lang Model LM-CR}
    
\addplot
    coordinates {(2000,4.7)(4000,6.3)(6000,7.1)(8000,7.6)(10000,7.4)(12000,7.2)(14000,6.9)(16000,6.6)(18000,6.5)(20000,6.2)};
    \addlegendentry{Con Lang Model NMT-CR}
    
\addplot
    coordinates {(2000,3.1)(4000,4)(6000,4.4)(8000,4.6)(10000,4.6)(12000,4.4)(14000,4.5)(16000,4.3)(18000,4.3)(20000,4.2)};
    \addlegendentry{Con Attn Lang Model LM-CR}
    
\addplot
    coordinates {(2000,5.1)(4000,6.3)(6000,6.9)(8000,6.9)(10000,7)(12000,6.8)(14000,6.7)(16000,6.5)(18000,6.4)(20000,6.2)};
    \addlegendentry{Con Attn Lang Model NMT-CR}
    
\addplot
    coordinates {(2000,12.8)(4000,13.3)(6000,12.9)(8000,12.1)(10000,11.4)(12000,10.7)(14000,10.2)(16000,9.8)(18000,9.5)(20000,9.2)};
    \addlegendentry{Con Bin Class Model LM-CR}
    
\addplot
    coordinates {(2000,13.3)(4000,13.7)(6000,13.3)(8000,12.4)(10000,11.6)(12000,11.1)(14000,10.5)(16000,10.2)(18000,9.8)(20000,9.5)};
    \addlegendentry{Con Bin Class Model NMT-CR}
    
\end{axis}
\end{tikzpicture}
\caption{Results on the Development Set - Depending on the highest F-score, the separation thresholds for the models are determined.}
\label{fig:f1_score}
\vspace*{15px}
\end{figure*}

\begin{table*}[!h]
\centering
\begin{tabular}{|l|llll|}
\hline 
Model & Perplexity & Precision & Recall & F-score\\
\hline 
Baseline Lang Model & 115 & 4.12\% & \textbf{15.42\%} & 6.51\%\\
Context Lang Model LM-CR & 79 & 5.18\% & 11.05\% & 7.06\%\\ 
Context Lang Model NMT-CR & \textbf{76} & \textbf{5.43\%} & 11.31\% & \textbf{7.34\%}\\
Context Attn Lang Model LM-CR & 125 & 3.05\% & 7.32\% & 4.31\% \\
Context Attn Lang Model NMT-CR & 82 & 4.68\% & 12.43\% & 6.8\% \\
\hline 
Baseline Bin Class Model & - & 8.94\% &  \textbf{11.76\%} & 10.16\%\\
Context Bin Class Model LM-CR & - & 13.4\% & 10.72\% & 11.92\%\\
Context Bin Class Model NMT-CR & \textbf{-} & \textbf{13.77\%} & 11.01\% & \textbf{12.23\%}\\
\hline 
\end{tabular}
\caption{Final Results on the Test-Set - The best performing models in the unsupervised and supervised category are bolded.}
\label{tab:final_result}
\end{table*}

To assess the final results, the F-score and perplexity values are evaluated on the development set. Figure \ref{fig:f1_score} shows the F-scores for separation thresholds in the range of $2,000$ to $20,000$ words for the best hyper-parameter settings of every model topology. The displayed model performances on the development-set already show a clear separation between the unsupervised and supervised models, with the supervised models reaching higher F-scores at lower thresholds. Depending on the individual graph, the optimal threshold for every model is independently determined and subsequently applied on the test set.

For the unsupervised models, the perplexity is additionally evaluated and taken into account when selecting the best model hyper-parameters such as the learning rate and the training epoch.

The best instance of every topology is evaluated on the mutually exclusive test set. The final results are shown in table \ref{tab:final_result}.

The unsupervised baseline language model, only taking the local context of a sentence into account, reaches a F-score of 6.51\% and a perplexity of 115 on the test-set. The second baseline system, the supervised binary classification model, achieves a F-score of 10.16\%. Based on these values for the unsupervised and supervised baselines, the context-aware systems are evaluated on the test-set. 

The unsupervised context-aware language model is evaluated with both sentence representations described in section \ref{training}: the language model context representation (\textit{Context Lang Model LM-CR}) and the NMT context embedding (\textit{Context Lang Model NMT-CR}). The two model instances reach F-scores of 7.06\% and 7.34\% respectively. Compared to the unsupervised baseline, the better performing NMT based context-aware model increases the F-score by 0.83\% absolute and 12.75\% relative. This model also reduces the perplexity compared to the baseline model, showing the general superiority of context-aware models for the word prediction task.

The two attention based language models (\textit{Context Attn Lang Model LM-CR} and \textit{Context Attn Lang Model NMT-CR}) are not able to enhance the performance compared to the previously described context-aware language models, especially due to the low precision values. Therefore, it is not advisable to add an attention mechanism.

Both supervised context-aware binary classification models (\textit{Context Bin Class Model LM-CR} and \textit{Context Bin Class Model NMT-CR}) enhance the performance of the supervised baseline. The NMT based sentence representation again outperforms the language model sentence embeddings, indicating the superiority of the NMT based sentence representation for all evaluated model topologies. Compared to the supervised baseline, the context component enhances the system by 2.07\% absolute and 20.37\% relative, increasing the F-score form 10.16\% to 12.23\%. The context component thereby primarily enhances the precision, while the recall remains mostly constant.

The best model topologies in the unsupervised and supervised category are bolded in the final performance comparison (See table \ref{tab:final_result})

\section{Conclusion}
In this paper, we present a novel architecture for context-aware computational models. To outline the potentials of the contextual approach, we compare the model to sentence-based baseline models showing the effects of a hierarchical context on state-of-the-art systems. With a unsupervised context extension applied to the sentence-based models, the performance improved by a factor of 12.75\% relative. Adding an additional attention mechanism to the model impairs the results and is therefore not advisable. Applying the context on a supervised model shows an enhancement of 20.37\% relative. 

The consistent improvements of the performance for unsupervised and supervised models illustrate the significance of an extended context for the task of semantic error detection. The result, hence, shows the importance of a context to semantic error detection.

With this work, the general superiority of context-aware system over sentence-based approaches to find semantic errors has been shown. To reduce the computational efforts, multiple constraints have been introduced, including a limited topology size and a confined vocabulary. To further enhance the performances of the context-aware models, an increased number of computational layers can be employed to learn more abstract relations on the data. Through the restricted vocabulary in this work, an extended or open vocabulary should be taken into account to further excel the quality of the model.

\bibliography{acl2018}

\begin{thebibliography}{19}
\expandafter\ifx\csname natexlab\endcsname\relax\def\natexlab#1{#1}\fi

\bibitem[{Bahdanau et~al.(2014)Bahdanau, Cho, and Bengio}]{attention}
Dzmitry Bahdanau, Kyunghyun Cho, and Yoshua Bengio. 2014.
\newblock Neural machine translation by jointly learning to align and
  translate.
\newblock \emph{CoRR}, abs/1409.0473.

\bibitem[{Bengio et~al.(2003)Bengio, andPascal Vincent, and Jauvin}]{bengio}
Yoshua Bengio, Réjean~Ducharme andPascal Vincent, and Christian Jauvin. 2003.
\newblock A neural probabilistic language model.
\newblock \emph{Journal of Machine Learning Research}, 3:1137?1155.

\bibitem[{Cettolo et~al.(2012)Cettolo, Girardi, and Federico}]{wit3}
Mauro Cettolo, Christian Girardi, and Marcello Federico. 2012.
\newblock Wit3: Web inventory of transcribed and translated talks.
\newblock \emph{Proceedings of the EAMT Conference}, 16:261--268.

\bibitem[{Chomsky(2002)}]{chomsky2002syntactic}
Noam Chomsky. 2002.
\newblock \emph{Syntactic structures}.
\newblock Walter de Gruyter.

\bibitem[{Chung et~al.(2016)Chung, Ahn, and Bengio}]{chung_et_al_16}
Junyoung Chung, Sungjin Ahn, and Yoshua Bengio. 2016.
\newblock \href {http://arxiv.org/abs/1609.01704} {Hierarchical multiscale
  recurrent neural networks}.
\newblock \emph{CoRR}, abs/1609.01704.

\bibitem[{De~Marneffe et~al.(2006)De~Marneffe, MacCartney, Manning
  et~al.}]{dependency_parse_tree}
Marie-Catherine De~Marneffe, Bill MacCartney, Christopher~D Manning, et~al.
  2006.
\newblock Generating typed dependency parses from phrase structure parses.
\newblock In \emph{Proceedings of LREC}, 2006, pages 449--454. Genoa Italy.

\bibitem[{Hochreiter and Schmidhuber(1997)}]{lstm}
Sepp Hochreiter and Juergen Schmidhuber. 1997.
\newblock Long short-term memory.
\newblock 9:1735--80.

\bibitem[{Hwang and Sung(2016)}]{hwang_et_al_16}
Kyuyeon Hwang and Wonyong Sung. 2016.
\newblock \href {http://arxiv.org/abs/1609.03777} {Character-level language
  modeling with hierarchical recurrent neural networks}.
\newblock \emph{CoRR}, abs/1609.03777.

\bibitem[{Iyyer et~al.(2014)Iyyer, Boyd-Graber, Claudino, Socher, and
  III}]{nlp_example_related_iyyer}
Mohit Iyyer, Jordan Boyd-Graber, Leonardo Claudino, Richard Socher, and
  Hal~Daume III. 2014.
\newblock A neural network for factoid question answering over paragraphs.
\newblock In \emph{Empirical Methods in Natural Language Processing}.

\bibitem[{Iyyer et~al.(2017)Iyyer, Yih, and Chang}]{iyyer2017search}
Mohit Iyyer, Wen-tau Yih, and Ming-Wei Chang. 2017.
\newblock Search-based neural structured learning for sequential question
  answering.
\newblock In \emph{Proceedings of the 55th Annual Meeting of the Association
  for Computational Linguistics (Volume 1: Long Papers)}, volume~1, pages
  1821--1831.

\bibitem[{Kingma and Ba(2014)}]{adam}
Diederik~P. Kingma and Jimmy Ba. 2014.
\newblock Adam: {A} method for stochastic optimization.
\newblock \emph{CoRR}, abs/1412.6980.

\bibitem[{Kumar et~al.(2015)Kumar, Irsoy, Su, Bradbury, English, Pierce,
  Ondruska, Gulrajani, and Socher}]{kumar_related}
Ankit Kumar, Ozan Irsoy, Jonathan Su, James Bradbury, Robert English, Brian
  Pierce, Peter Ondruska, Ishaan Gulrajani, and Richard Socher. 2015.
\newblock Ask me anything: Dynamic memory networks for natural language
  processing.

\bibitem[{Lashley(1951)}]{lashley1951problem}
Karl~Spencer Lashley. 1951.
\newblock The problem of serial order in behavior.
\newblock In \emph{Cerebral mechanisms in behavior}, pages 112--136.

\bibitem[{Mikolov et~al.(2010)Mikolov, Karafi{\'a}t, Burget, Cernock{\`y}, and
  Khudanpur}]{rnn_basics}
Tomas Mikolov, Martin Karafi{\'a}t, Lukas Burget, Jan Cernock{\`y}, and Sanjeev
  Khudanpur. 2010.
\newblock Recurrent neural network based language model.
\newblock In \emph{Interspeech}, volume~2, page~3.

\bibitem[{Mikolov et~al.(2013)Mikolov, Sutskever, Chen, Corrado, and
  Dean}]{seq2seq_with_nn}
Tomas Mikolov, Ilya Sutskever, Kai Chen, Greg Corrado, and Jeffrey Dean. 2013.
\newblock Distributed representations of words and phrases and their
  compositionality.

\bibitem[{Newell et~al.(1958)Newell, Shaw, and Simon}]{newell1958elements}
Allen Newell, John~Calman Shaw, and Herbert~A Simon. 1958.
\newblock Elements of a theory of human problem solving.
\newblock \emph{Psychological review}, 65(3):151.

\bibitem[{Paperno et~al.(2016)Paperno, Kruszewski, Lazaridou, Pham, Bernardi,
  Pezzelle, Baroni, Boleda, and Fern{\'{a}}ndez}]{Paperno-et-al-16}
Denis Paperno, Germ{\'{a}}n Kruszewski, Angeliki Lazaridou, Quan~Ngoc Pham,
  Raffaella Bernardi, Sandro Pezzelle, Marco Baroni, Gemma Boleda, and Raquel
  Fern{\'{a}}ndez. 2016.
\newblock \href {http://arxiv.org/abs/1606.06031} {The {LAMBADA} dataset: Word
  prediction requiring a broad discourse context}.
\newblock \emph{CoRR}, abs/1606.06031.

\bibitem[{Rosenbaum et~al.(1983)Rosenbaum, Kenny, and
  Derr}]{rosenbaum1983hierarchical}
David~A Rosenbaum, Sandra~B Kenny, and Marcia~A Derr. 1983.
\newblock Hierarchical control of rapid movement sequences.
\newblock \emph{Journal of Experimental Psychology: Human Perception and
  Performance}, 9(1):86.

\bibitem[{Wang et~al.(2017)Wang, Tu, Way, and Liu}]{translation_hierarchical}
Longyue Wang, Zhaopeng Tu, Andy Way, and Qun Liu. 2017.
\newblock Exploiting cross-sentence context for neural machine translation.

\end{thebibliography}
\bibliographystyle{acl_natbib}

\end{document}